\documentclass[sigconf]{acmart}

\usepackage{amsmath,amsfonts}
\usepackage{array}
\usepackage[caption=false,font=normalsize,labelfont=sf,textfont=sf]{subfig}
\usepackage{textcomp}
\usepackage{stfloats}
\usepackage{url}
\usepackage{verbatim}
\usepackage{graphicx}
\usepackage{hyperref}
\usepackage{amsfonts}
\usepackage{bbding}
\usepackage{graphicx}

\usepackage{amssymb}
\usepackage{booktabs}
\usepackage{threeparttable}
\usepackage{enumitem}

\usepackage{amsthm}

\usepackage{algorithm}
\usepackage{algorithmic}

\usepackage{newfloat}
\usepackage{listings}
\DeclareCaptionStyle{ruled}{labelfont=normalfont,labelsep=colon,strut=off} 
\floatstyle{ruled}
\newfloat{listing}{tb}{lst}{}
\floatname{listing}{Listing}

\usepackage{multirow}
\usepackage{color,xcolor}
\usepackage{colortbl}

\AtBeginDocument{%
  }

\copyrightyear{2023}
\acmYear{2023}
\setcopyright{acmlicensed}\acmConference[MM '23]{Proceedings of the 31st ACM International Conference on Multimedia}{October 29-November 3, 2023}{Ottawa, ON, Canada}
\acmBooktitle{Proceedings of the 31st ACM International Conference on Multimedia (MM '23), October 29-November 3, 2023, Ottawa, ON, Canada}
\acmPrice{15.00}
\acmDOI{10.1145/3581783.3611698}
\acmISBN{979-8-4007-0108-5/23/10}
\begin{document}

\title{Orthogonal Uncertainty Representation of Data Manifold for Robust Long-Tailed Learning}


\author{Yanbiao Ma}
\affiliation{%
  \institution{Xidian University}
  \city{Xi'an}
  \country{China}}
\email{ybmamail@stu.xidian.edu.cn}

\author{Licheng Jiao}
\authornote{Corresponding author(s).}
\affiliation{%
  \institution{Xidian University}
  \city{Xi'an}
  \country{China}}
\email{lchjiao@mail.xidian.edu.cn}

\author{Fang Liu}
\affiliation{%
  \institution{Xidian University}
  \city{Xi'an}
  \country{China}}
\email{f63liu@163.com}

\author{Shuyuan Yang}
\affiliation{%
  \institution{Xidian University}
  \city{Xi'an}
  \country{China}}
\email{syyang@xidian.edu.cn}

\author{Xu Liu}
\affiliation{%
  \institution{Xidian University}
  \city{Xi'an}
  \country{China}}
\email{xuliu361@163.com}

\author{Lingling Li}
\affiliation{%
  \institution{Xidian University}
  \city{Xi'an}
  \country{China}}
\email{llli@xidian.edu.cn}

\renewcommand{\shortauthors}{Yanbiao Ma et al.}

\begin{abstract}
In scenarios with long-tailed distributions, the model's ability to identify tail classes is limited due to the under-representation of tail samples. Class rebalancing, information augmentation, and other techniques have been proposed to facilitate models to learn the potential distribution of tail classes. The disadvantage is that these methods generally pursue models with balanced class accuracy on the data manifold, while ignoring the ability of the model to resist interference. By constructing noisy data manifold, we found that the robustness of models trained on unbalanced data has a long-tail phenomenon. That is, even if the class accuracy is balanced on the data domain, it still has bias on the noisy data manifold. However, existing methods cannot effectively mitigate the above phenomenon, which makes the model vulnerable in long-tailed scenarios. In this work, we propose an Orthogonal Uncertainty Representation (\emph{OUR}) of feature embedding and an end-to-end training strategy to improve the long-tail phenomenon of model robustness. As a general enhancement tool, \emph{OUR} has excellent compatibility with other methods and does not require additional data generation, ensuring fast and efficient training. Comprehensive evaluations on long-tailed datasets show that our method significantly improves the long-tail phenomenon of robustness, bringing consistent performance gains to other long-tailed learning methods.
\end{abstract}



\maketitle

\section{Introduction}\label{sec1}

Long-tailed recognition is an important challenge in computer vision, manifested by models trained on long-tailed data that tend to perform poorly for classes with few samples. Previous research has attributed this phenomenon
to the fact that the few samples in the tail classes do not well represent their true distribution, resulting in a shift
between the test and training domains \cite{paper5,paper41,paper55}. Numerous methods have been proposed to mitigate model bias. Class rebalancing \cite{paper2,paper3,paper8,paper15,paper20,paper21,paper23,paper29,paper34,paper37,paper43,paper45,paper51,paper54,paper57,paper58,paper62}, for example, aims to boost the weight of losses arising from tail classes, thereby pushing the decision boundary away from the tail class and improving the probability of correctly classifying the underlying distribution. Information augmentation \cite{paper5,paper9,paper13,paper18,paper22,paper24,paper26,paper32,paper33,paper42,paper50,paper53,paper54,paper60}, on the other hand, expands the observed distribution of the tail classes by introducing prior knowledge to facilitate the model learning of the underlying distribution. It is important to note that these methods default to the model being able to learn adequately and fairly at least for the samples in the training domain. However, we find that even if a model has balanced class accuracy over the training domain, its robustness still exhibits a long-tailed distribution, and existing methods do not improve the phenomenon well.

\begin{figure*}[!t]
\centering
\includegraphics[width=7in]{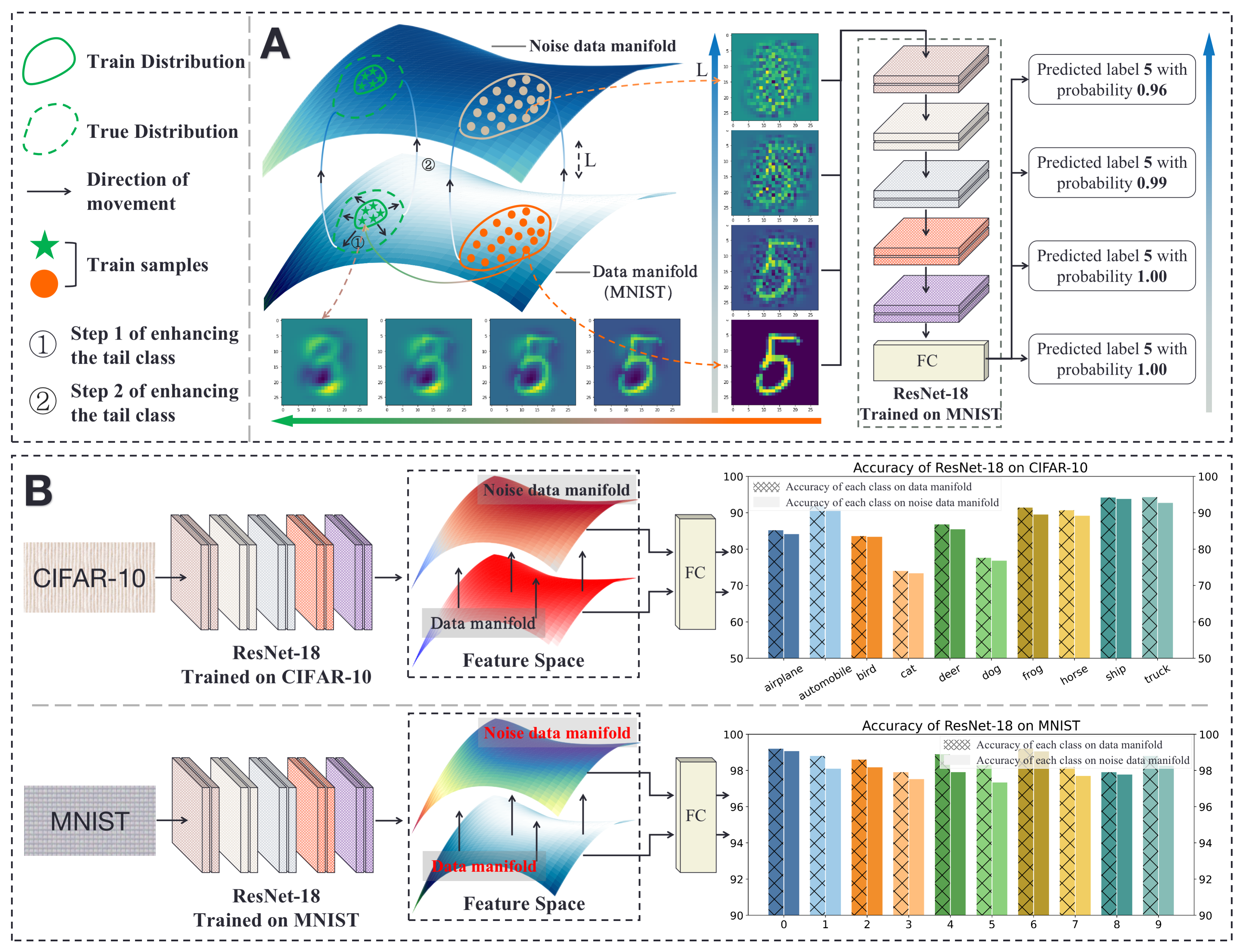}
\vskip -0.15in
\caption{A: Moving in the orthogonal direction along the data manifold produces a series of images where the level of noise increases with distance. Moving on the data manifold corresponds to successive changes in the different classes of samples. A trained deep neural network can predict
samples on noisy data stream shapes with a high confidence level. B: The network trained on the balanced dataset has excellent robustness. ResNet-18 was trained on CIFAR-10 and MNIST respectively and then tested for its performance on noisy data manifolds.}
\label{fig1}
\vskip -0.05in
\end{figure*}

Recent study \cite{paper11} indicates that moving in the direction orthogonal to the data manifold produces a series of noisy data manifolds, and the samples on these noisy data manifolds are noisy versions of the real samples. We construct the noisy data manifold corresponding to the training samples (i.e. the data manifold) on sample-balanced MNIST and CIFAR-10. It is found that ResNet-18 trained on the data manifold can correctly recognize noisy samples from all classes with high confidence (Fig.\ref{fig1}A, even images that are meaningless to the human eye) and that the class accuracy on the noisy data manifold is balanced (Fig.\ref{fig1}B). The same experiments are then performed on the CIFAR-10-LT. Unexpectedly, we find that although ResNet-18 performs well and fairly for each class on the CIFAR-10-LT training set, the accuracy of the model for the tail class decreases rapidly as the distance between the noisy data manifold and the data manifold increases, leading to a long-tailed distribution of model robustness (Fig.\ref{fig3}A). This suggests that the decision
surface is not only biased towards the tail class in terms of data manifold, but that the degree of bias towards the tail class increases with increasing distance between the noisy data manifold and the data manifold (Fig.\ref{fig3}B).

The long-tailed phenomenon of robustness makes the model more vulnerable in test scenarios. To alleviate this phenomenon, we propose the orthogonal uncertainty representation of tail classes in the feature space. It is well-compatible with existing methods to improve the performance of the model on both the underlying distribution and the noisy data manifold. The main contributions of this work are summarised as follows.

\vspace{-4pt}
\begin{itemize}[labelsep = 5 pt, leftmargin = 13pt]
\item[1)]We discover and define the long-tailed phenomenon of model robustness on unbalanced datasets and propose a corresponding measure of unbalance, \emph{RIF}. (Section \ref{sec3})
\item[2)]We propose the orthogonal uncertainty representation (\emph{OUR}) of feature embedding. \emph{OUR} is simple and efficient, plug and-play, and does not affect the speed of inference. (Section \ref{sec4.1})
\item[3)]We solve the problem that calculating the orthogonal direction of feature manifolds interrupts training and consumes time and video memory, enabling end-to-end and low-cost applications \emph{OUR}. (Section \ref{sec4.2})
\item[4)]Comprehensive experiments show that our method has excellent compatibility and generality, demonstrating superior performance on multiple long-tailed datasets and effectively improving the long-tailed phenomenon of model robustness.
\end{itemize}

\section{MOTIVATION: THE LONE-TAILED PHENOMENON OF MODEL ROBUSTNESS}\label{sec3}

In this section, we first introduce the method of constructing noisy data manifolds, and then discover and define the long-tail phenomenon of model robustness and the measure of imbalance factor. Finally, we analyze the factors that affect the performance of the tail class, thus pointing out the directions and goals of the research.

\subsection{Data manifold and noise data manifold}\label{sec3.1}
\subsubsection{Constructing noisy data manifold}\label{sec3.1.1}

The manifold distribution law \cite{paper19} considers that natural images distribute around a low-dimensional manifold in a high-dimensional space, called a data manifold. As shown in Fig.\ref{fig1}A, \cite{paper11} found that moving the sample points (i.e., images) along the direction orthogonal to the data manifold, the noise of the images continues to increase, and these sample points in the orthogonal direction constitute the noisy data manifold. The following describes how to generate the noisy data manifold.

Given an image dataset with the number of samples $N$, assume that the size of the image is $l \times w \times h = d$ and all samples are denoted as $X=[x_1,\dots,x_N]\in \mathbb{R}^{d\times N}$. In the d-dimensional sample space, each image is considered a point, and the set of points corresponding to all images constitutes the data manifold. The intrinsic dimension of a data manifold is usually smaller than the dimension $d$ of the linear space, so a direction vector $U\in \mathbb{R}^d$ orthogonal to the data manifold can be found to construct the noisy data manifold. If $U$ is strictly orthogonal to the data manifold, then its inner product with any vector $(x_i-c)\in \mathbb{R}^d, i=1,\dots,N$ is $0$, where $c=\frac{1}{N} {\textstyle \sum_{i=1}^{N}x_i}  $. Therefore, we solve for $U$ by optimizing the following objective.
\begin{equation}
\begin{split}
\min\sum_{i=1}^{N}((x_i-c)^TU)^2.
\end{split}
\end{equation}
Let $y_i=x_i-c\in \mathbb{R}^d$, then the optimization objective is transformed into
\begin{equation}
\begin{split}
\min\sum_{i=1}^{N}(y_i^TU)^2 & =\min\sum_{i=1}^{N}U^Ty_iy_i^TU=\min(U^T(\sum_{i=1}^{N}y_iy_i^T)U).
\end{split}
\end{equation}
Let $Y\! \! = \! [y_1,\dots,y_N] \! \in \! \mathbb{R}^{d\times N}$ and $ {\textstyle \sum_{i=1}^{N}y_iy_i^T=YY^T\in \mathbb{R}^{d\times d}} $. The optimization objective can be equivalent to
\begin{equation}
\begin{split}
\min(U^TYY^TU),\\
s.t. U^TU=1.
\end{split}
\label{equ3}
\end{equation}

Construct the Lagrangian function $L(U,\lambda)=U^TYY^TU-\lambda (U^TU-1)$, where $\lambda$ is a coefficient. By making $\frac{\partial L(U,\lambda)}{\partial U} $ and $\frac{\partial L(U,\lambda)}{\partial \lambda} $ equal to $0$,  respectively, we get
\begin{equation}
\begin{split}
YY^TU=\lambda U, \\
U^TU=1.
\end{split}
\end{equation}
Obviously, $YY^T$ is the covariance matrix of $X$ and $U$ is the eigenvector of $YY^T$. Further, from $(YY^TU,U)=(\lambda U,U)$ we can get
\begin{equation}
\begin{split}
\lambda = (YY^TU,U)=U^T(YY^T)^TU=U^TYY^TU.
\end{split}
\end{equation}

Therefore, in combination with equation (\ref{equ3}), the optimization objective is ultimately equivalent to $\min_U(\lambda)$. The above results show that $U$ can take the eigenvector corresponding to the smallest eigenvalue of $YY^T$.

\begin{figure}[h]
\centering
\includegraphics[width=20pc]{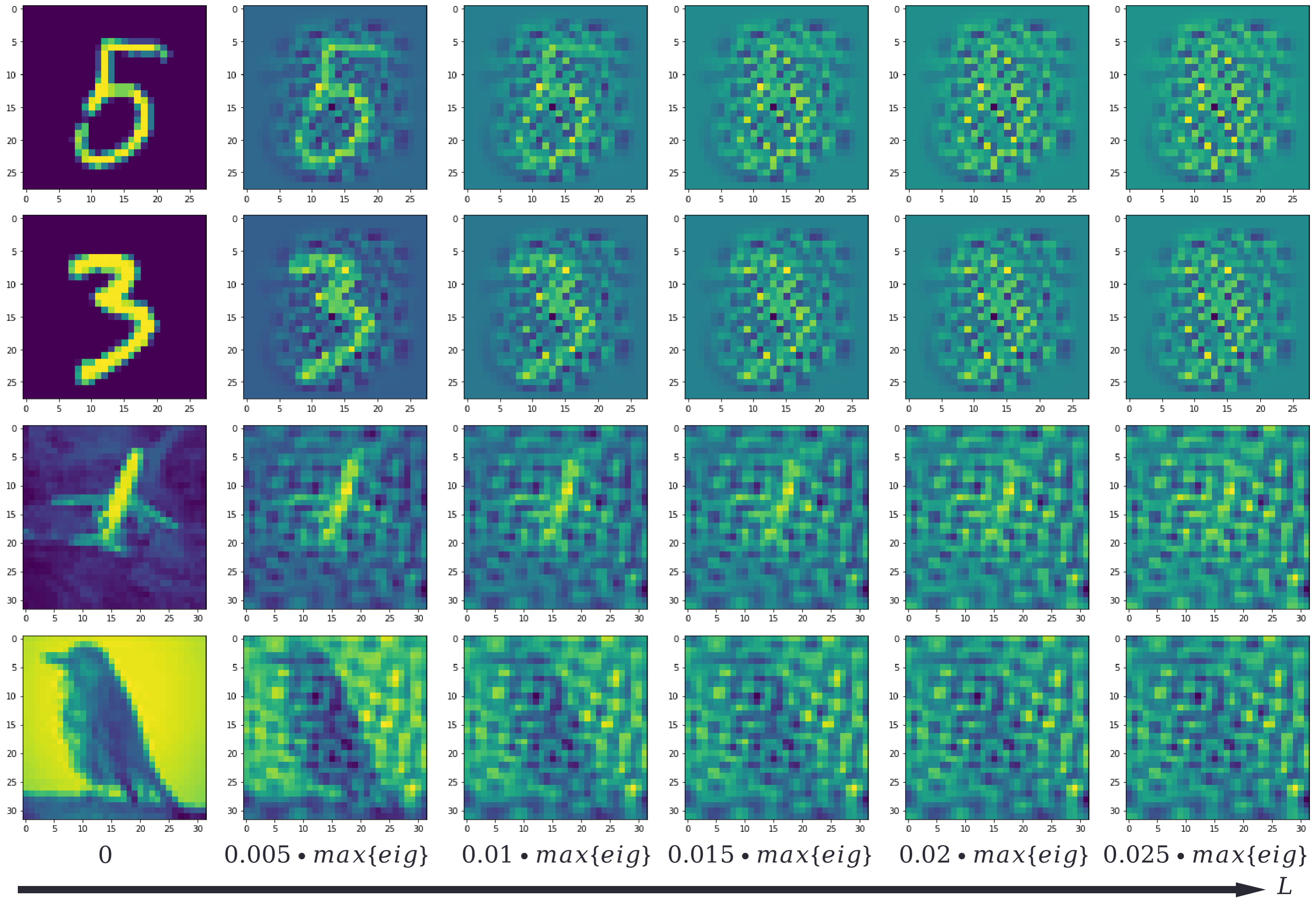}
\vskip -0.08in
\caption{The above two rows show the variation process of the samples in the MNIST dataset moving along $U$. The lower two rows show the variation process of the airplane and bird images in CIFAR-10 moving along $U$. $\max\{eig\}$ denotes the maximum eigenvalue of $YY^T$. It can be observed that when $L = 0.02 \times \max\{eig\}$, it is already difficult for the human eye to distinguish these images.}
\label{fig2}
\vskip -0.05in
\end{figure}

As shown in Fig.\ref{fig1}A, the data manifold formed by the sample set $X=[x_1,\dots,x_N]\in \mathbb{R}^{d\times N}$ is shifted by a distance $L$ along the direction $U$ to obtain the sample set $X'=X+LU$, and $X'$ forms a noise data manifold. The farther the noisy data manifold is from the data manifold, the noisier the image in $X'$ is. We correlate the value of $L$ with the maximum eigenvalue of $YY^T$. Studies on MNIST and CIFAR-10 show that images on noisy data manifold are almost unrecognizable by the human eye when $L$ is about $2\%$ of the maximum eigenvalue of $YY^T$ (Fig.\ref{fig2}). In the following, $L$ defaults to $2\%$ of the largest eigenvalue of the sample covariance matrix if not otherwise specified.

\begin{figure*}[t]
\centering
\includegraphics[width=17.6cm,height=10.1cm]{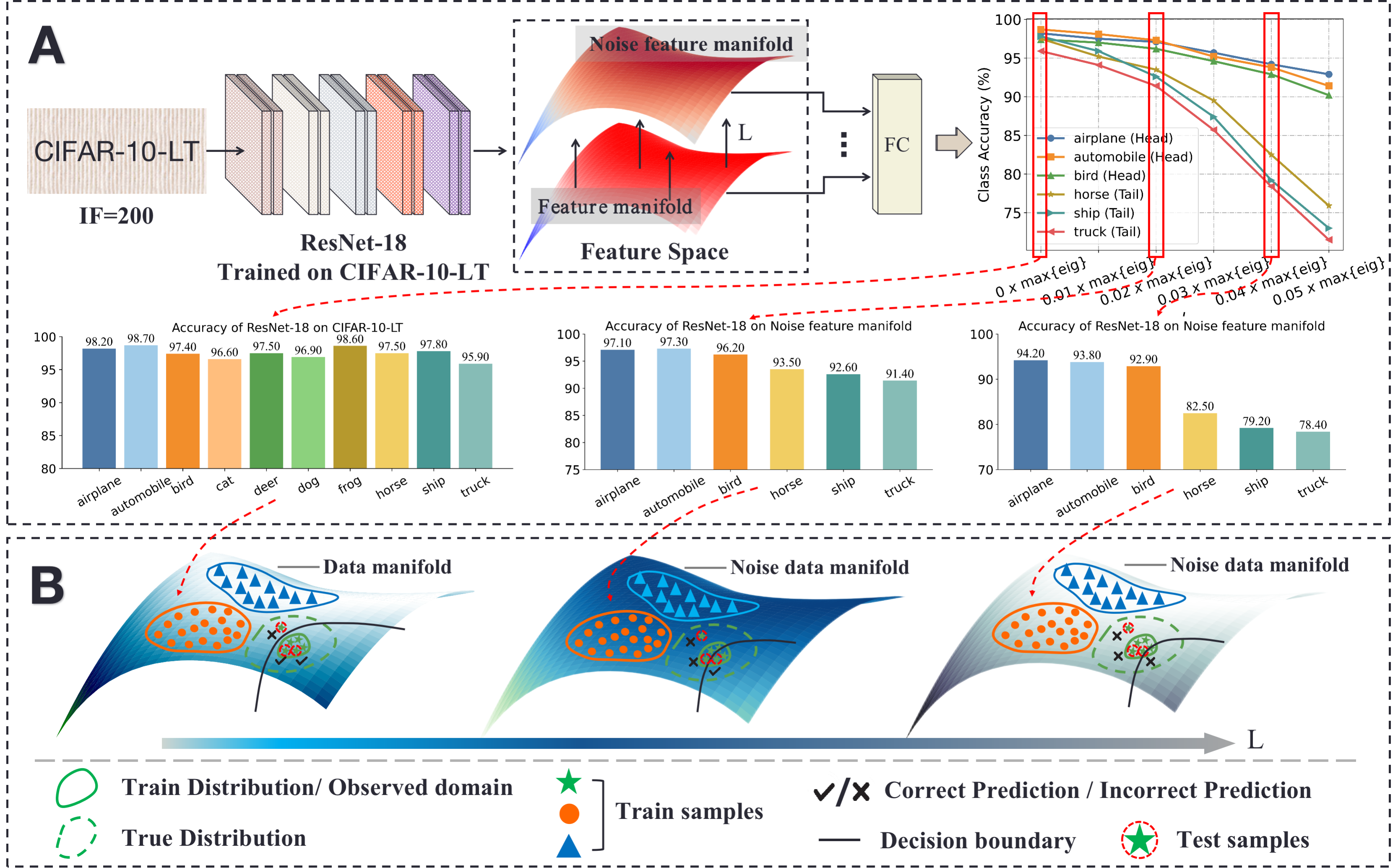}
\vskip -0.05in
\caption{A: Long-tailed phenomenon for robustness of ResNet-18 trained on CIFAR-10-LT. The \emph{RIF} increases as $L$ increases. B: When the test sample is outside the observed domain, the model may predict incorrectly. Combined with the trend in \emph{RIF}, we speculate that as $L$ increases, the decision surface on the noisy data manifold becomes more biased towards the tail class.}
\label{fig3}
\vskip -0.05in
\end{figure*}

\subsubsection{Why the orthogonal direction?}\label{sec3.1.2}

Moving a sample in a non-orthogonal direction may imply changes in the main features of that sample, \textbf{resulting in a conflict between the moved sample and the label}. For example, in Fig.\ref{fig1}A, moving from the orange sample to the green sample on the data manifold corresponds to a gradual change in the image from the number $5$ to the number $3$, while the label of the orange sample is kept constant. Of course, shifts in non-orthogonal directions may be a potential method for generating adversarial samples, which we do not discuss in depth in this work. Since the direction of random noise is uncontrollable, it cannot be used to produce the noisy data manifold. In Section \ref{sec5} we compare noise-based data augmentation with our method.

\subsection{Equity in model robustness on balanced data}\label{sec3.2}

Surprisingly, Fig.\ref{fig1}A shows that ResNet-18 trained on MNIST can correctly classify samples on noisy data manifolds with high confidence even in the face of heavily noisy images that are meaningless to the human eye. ResNet-18 has the same setup as ResNet-32 in section \ref{sec5.1}.

We explore and find that the above phenomenon also exists in the feature space. First, ResNet-18 is trained on sample-balanced MNIST and CIFAR-10 respectively, and extracts the features of all samples, which constitute the feature data manifolds (referred to as feature manifolds below). We test the classification accuracy of ResNet-18 on the feature manifolds as well as the noisy feature manifolds respectively, and the experimental results are shown in Fig.\ref{fig1}B. ResNet-18 has excellent generalization performance for each class on noisy feature manifolds, and the overall test accuracy on noisy feature manifolds for CIFAR-10 and MNIST is only $1.04\%$ and $0.49\%$ lower than the overall test accuracy on feature manifolds, respectively.

The above experiments combined show that a model trained on a sample-balanced data manifold, whether in image space or feature space, typically has fair and well robustness on noisy data manifolds, and is far superior to humans. What makes us curious is whether the above phenomenon also exists in long-tailed data. From the sample level, the noise generalization ability of the model trained on the balanced dataset is relatively balanced for each sample. Does the same phenomenon exist in the long-tailed data?

\vspace{-2pt}
\subsection{Inequities in model robustness on long-tailed data}\label{sec3.3}

First think about a question: when training a classification model on a long-tailed dataset, without considering the performance of the model outside the training domain, does the model learn equally well on the training set for both head and tail class samples? We trained ResNet-18 on CIFAR-10-LT (imbalance factor = $200$) and show the training accuracy of ResNet-18 on all classes in Fig.\ref{fig3}A. As can be seen, ResNet-18 can recognize training samples for both head and tail classes with high accuracy, which indicates that ResNet-18 adequately fits the distribution of the training domain. However, is this sufficient to prove that the model learns fairly for each class over the training domain?

We further explore the ability of the model to generalize over noisy data manifolds in the feature space. First, training ResNet-18 on CIFAR-10-LT and extracting the features of all training samples, the maximum eigenvalue of the sample covariance matrix is denoted as $\max\{eig\}$. Then the shift distance was gradually increased along the orthogonal direction of the feature manifold until $L = 0.05 \times \max\{eig\}$, and multiple noisy feature manifolds were generated in this process. Testing the classification performance of ResNet-18 on each noisy feature manifold, the experimental results are shown in Fig.\ref{fig3}A. We find that the performance of the tail classes decreases earlier and faster than the head classes as the distance $L$ increases, resulting in class accuracies on the noisy feature manifold that exhibit increasingly extreme long-tailed distributions. \textbf{This suggests that the robustness of models trained on long-tailed datasets to classes also exhibits a long-tailed distribution}. Not only is the decision surface skewed toward the tail class on the feature manifold, but it is also skewed even more heavily toward the tail class in the noisy feature manifold (see Fig.\ref{fig3}B). In the following, we formally define the long-tailed phenomenon of model robustness and its imbalance metric.

\begin{definition}[\textbf{The long-tailed phenomenon of model robustness}]
Given a dataset containing $C$ classes (data manifold) and training a classification model on it, test the class accuracy $A_1,\dots, A_C$ of the model on the data manifold. Then construct a noisy data manifold and test the class accuracy $A'_1,\dots, A'_C$ of  the classification model on it. The difference in class accuracy, $A_i-A'_i,i=1,\dots,C$, reflects the robustness of the model to the classes. The long-tailed phenomenon of model robustness arises when the difference in the accuracy of the classes is unbalanced.
\end{definition}

\begin{definition}[\textbf{Imbalance factor for model robustness}]
The imbalance factor for model robustness is defined as 
\begin{small}
\begin{equation}
\begin{split}
RIF = \max\{A_i-A'_i\}-\min\{A_i-A'_i\}(i=1,\dots,C).
\nonumber
\end{split}
\end{equation}
\end{small}
A larger RIF indicates that the model is more imbalanced in its robustness to the class. When the robustness is balanced, RIF = $0$.
\end{definition}

Even when fully fitting the training domain, the model is not as robust to the tail classes as the head classes, and this unfair performance may be even worse on the test set. The above results lead us to consider more comprehensively the reasons affecting the performance of the tail class.

\subsection{Rethinking the factors affecting tail class performance in long-tailed recognition}\label{sec3.4}

First, we define the meaning of two concepts. For a class, the observed distribution is the distribution consisting of the available samples, and the underlying distribution is the true distribution in addition to the observed distribution. Combining with the discoveries in Section \ref{sec3.3}, we believe that the reasons for the performance limitations of the tail class are as follows.

\begin{itemize}
\item[1)] As shown in Fig.\ref{fig3}B, the few samples of the tail classes do not well represent its true distribution, so the performance of the model is limited outside the training domain \cite{paper5,paper41}. How to recover the underlying distribution of the tail classes is the key to the study.
\item[2)] The robustness of the model to the classes shows a long-tailed distribution. As illustrated in Fig.\ref{fig3}B, the decision boundary of the model trained on long-tailed data is not only biased towards the tail class in the data manifold but also biased more severely in the noisy data manifold. This limits the noise invariance of the model on tail classes.
\end{itemize}

Previous studies have not taken into account the second cause of damage to the tail class, and simply expanding the data on the data manifold is not sufficient to mitigate the severe bias of the decision surface on the noisy data manifold. In this work, we propose a simple and efficient method with good compatibility to compensate for the shortcomings of existing methods.


\begin{figure*}[t]
\centering
\includegraphics[width=7in]{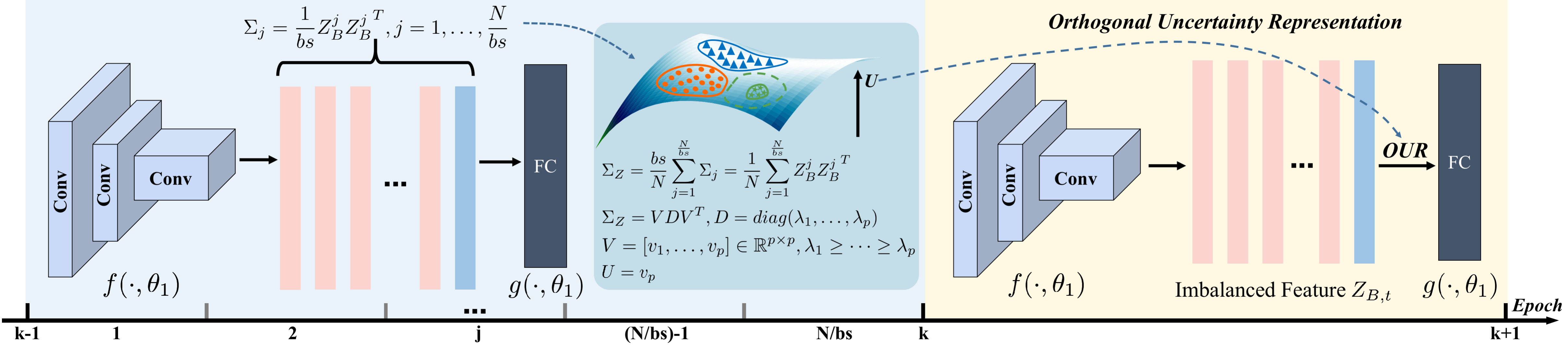}
\vskip -0.13in
\caption{End-to-end training process with \emph{OUR}. When applying \emph{OUR} in the $k$-th training epoch, the covariance matrix corresponding to each batch feature needs to be calculated and saved in the $k-1$-th epoch. At the end of the $k-1$-th epoch, the saved N/bs covariance matrices are used to calculate the corresponding feature covariance matrix of the entire dataset, which further yields the orthogonal direction $U$ of the feature manifold and is utilized in the $k$-th epoch.}
\label{fig4}
\vskip -0.05in
\end{figure*}

\section{ROBUST DEEP LONG-TAILED LEARNING}\label{sec4}
Consistent with \cite{paper1,paper13,paper19,paper22}, we are interested in ways to enhance tail classes in the feature space. In the following, the data manifold in the feature space is called the feature manifold, and the corresponding noisy data manifold is called the noisy feature manifold. To mitigate the long tail phenomenon of model robustness, we propose the orthogonal uncertainty representation (\emph{OUR}) of features in Section \ref{sec4.1}. In Section \ref{sec4.2}, we propose an end-to-end training scheme that substantially reduces the time cost and memory consumption of applying \emph{OUR}.

\subsection{Orthogonal uncertainty representation}\label{sec4.1}

The orthogonal uncertainty representation aims to augment the tail class samples along the orthogonal direction of the feature manifold. Given a long-tailed dataset $X$ containing $C$ classes and a deep neural network $Model=\{f(x,\theta_1),g(z,\theta_2)\}$, where $f(x,\theta_1)$ denotes a feature sub-network with parameter $\theta_1$ and $g(z,\theta_2)$ denotes a classifier with parameter $\theta_2$. The feature embedding corresponding to $X$ is assumed to be $Z\! =\! f(X,\theta_1)\! \!=\! \! [z_1,\dots,z_N] \! \! \in \! \! \mathbb{R}^{p\times N}$, where $N= {\textstyle \sum_{i=1}^{C}}N_i $, $p$ is the sample dimension and $N_i$ denotes the sample number of class $i$. $Z$ forms a feature manifold and calculates the sample covariance matrix $\Sigma_Z\! \! =\! \! \frac{1}{N}ZZ^T \! \! \in \mathbb{R}^{p\times p}$. The maximum and minimum eigenvalues of $\Sigma_Z$ are denoted by $\lambda_{\max}$ and $\lambda_{\min}$, respectively. The orthogonal direction $U\! \in \! \mathbb{R}^p$ of the feature manifold is the eigenvector corresponding to $\lambda_{\min}$.

Suppose a batch of samples is encoded by $f(X_B,\theta_1)$ as $Z_B\in \mathbb{R}^{p\times bs}$ and $bs$ as the batch size, where the feature embedding belonging to tail class $t$ is $Z_{B,t}=[z_t^1,\dots,z_t^{n_t}]\in \mathbb{R}^{p\times n_t}$. We model the uncertainty representation of features by applying perturbations along the direction $U$ for $Z_{B,t}$, thereby enhancing the noise invariance of the model to the tail class $t$ on noisy data manifolds. The specific form can be formulated as
\begin{equation}
\begin{split}
OUR(Z_{B,t})&=\! \! \! \! \! \!  \overbrace{Z_{B,t}+\mu \lambda_{mean}[\varepsilon_1U,\dots,\varepsilon_{n_t}U]\in \mathbb{R}^{p\times n_t}}^{Orthogonal \hspace{0.7mm} Uncertainty \hspace{0.7mm} Representation \hspace{0.7mm} of \hspace{0.7mm} Z_{B,t}}, \\ 
\varepsilon_i&\sim N(0,1),i=1,\dots,n_t.
\end{split}
\end{equation}

$\varepsilon_1,\dots,\varepsilon_{n_t}$ all follow a standard Gaussian distribution and are independent of each other, they increase the uncertainty of each feature embedding in $Z_{B,t}$. $\lambda_{mean}$ denotes the average of the top $10$ eigenvalues. Our study in Section \ref{sec3} shows that images on noisy feature manifolds are no longer recognizable to the human eye when the distance to the feature manifold is $0.02\lambda_{max}$. To improve the stability of \emph{OUR}, we use $\lambda_{mean}$ instead of $\lambda_{max}$ to perturb the features. The $\mu \lambda_{mean}$ term therefore guarantees a reasonable range of perturbations and the reasonable choice of $\mu$ will be discussed in Section \ref{sec5.2}. However, an additional consideration is that calculating the orthogonal direction $U$ of the feature manifold interrupts training. We detail the difficulties faced in practice in the next subsection.

\subsection{End-to-end training with \emph{OUR}}\label{sec4.2}
The parameters of the model change continuously with training, resulting in an offset between features extracted from the same sample at different periods. Therefore, when applying the orthogonal uncertainty representation (\emph{OUR}) in the feature space, the orthogonal direction of the feature manifold needs to be continuously updated. However, calculating the orthogonal direction of the feature manifold requires re-extracting features from the entire dataset, which significantly increases the time cost and interrupts the training, complicating the training process.

The feature slow shift phenomenon \cite{paper54} indicates that as the training epoch increases, the shift between the historical and latest features of the same sample decreases to the point where the historical features can be used to approximate the latest features. Assume that \emph{OUR} is applied from the $k$-th epoch. Given that the entire dataset can be traversed in a single training epoch, the intuitive solution is to save all batches of features from the $k-1$-th training epoch in place of the latest features $Z$ for the entire dataset. Before the $k$-th epoch, the covariance matrix is calculated with the saved features, and then the orthogonal direction $U$ of the feature manifold is further obtained. And so on, the historical features saved in the $k$-th epoch will be used to calculate the orthogonal direction used in the $k+1$-th epoch. \cite{paper7,paper54,paper55} shows that in the training of classification models, only $5$ epochs are usually needed to make the shift of features small enough. And $k$ is significantly larger than $5$, so there is no need to be concerned about the feature shift not being small enough.

Although the above approach avoids extracting features from the entire dataset, additional storage space is still required to hold all the features generated in an epoch. To further reduce memory consumption, we propose to calculate and save the covariance matrix $\Sigma_j=\frac{1}{bs}Z_B^j {Z_B^j}^T, j=1,\dots,\frac{N}{bs}$ for each batch of features in the $k$-th epoch, where $Z_B^j\in \mathbb{R}^{p\times bs}$ denotes the $j$-th batch of features. The sum of $\Sigma_1,\dots,\Sigma_{\frac{N}{bs}}$ is then used to approximate the covariance matrix of all features from the dataset. Specifically, when the $k$-th epoch ends, the feature covariance matrix of the entire dataset can be approximated as
\vspace{-2pt}
\begin{equation}
\begin{split}
\Sigma_Z=\frac{bs}{N}\sum_{j=1}^{\frac{N}{bs}}\Sigma_j=\frac{bs}{N}(\frac{1}{bs}Z_B^1{Z_B^1}^T+\cdots+ \\ 
\frac{1}{bs}Z_B^{\frac{N}{bs}}{Z_B^{\frac{N}{bs}}}^T )=\frac{1}{N}\sum_{j=1}^{\frac{N}{bs}}Z_B^j{Z_B^j}^T\in \mathbb{R}^{p\times p}  
\nonumber
\end{split}
\end{equation}

\vspace{-2pt}
Calculate the orthogonal direction $U$ of the feature manifold based on $\Sigma_Z$ and apply it to the next training epoch. Fig.\ref{fig4} illustrates the process of applying \emph{OUR} end-to-end. At the cost of negligible memory consumption (only $\frac{N}{bs}$ covariance matrices need to be stored), we solve the problem of interrupted training and time consumption when calculating the orthogonal direction of the feature manifold.

Since \emph{OUR} is dedicated to mitigating the long-tailed phenomenon of robustness found for the first time, it has good compatibility and generality as it does not overlap with the research aims of other methods. Our experiments also show that existing methods are not effective in mitigating the long-tailed phenomenon of robustness (Fig.\ref{fig6}). Listing \ref{lst1} demonstrates a simple implementation of \emph{OUR} that can easily be combined with existing methods. Fig.\ref{fig1}A illustrates the two steps in enhancing the tail classes. We hope to co-train to improve the model's performance on both underlying distribution and noisy data manifold.

\begin{listing}[h]%
\caption{End-to-end training with OUR}%
\label{lst1}%
\begin{lstlisting}[language=Python]
for epoch in range(M):
    Q = np.empty([N/bs, p, p]) # bs is batch size 
    # X_B: data, y_B: labels
    for X_B, y_B, j in loder(N/bs):
        Z_B = f(X_B, theta_1)
        if epoch == k-1:
            Sigma = np.matmul(Z_B, Z_B.T)
            Q[j] = Sigma
        elif epoch >= k:
            Sigma = np.matmul(Z_B, Z_B.T)
            Q[j] = Sigma
            for i in range(C):
            # Execute OUR on the tail category
                if Z_B_i is a tail category:
                    Z_B_i = OUR(Z_B_i, U, mu)
        y^  = g(Z_B, theta_2)
        loss = loss function(y_B, y^)
        loss.backward()
        optimizer.step()
    Sigma_Z = np.sum(Q, axis = 0)/N
    vals, vecs = np.linalg.eig(Sigma_Z)
    U = vecs[:, p-1]
    labmda_mean = np.mean(vecs[:, 0:10])
\end{lstlisting}
\end{listing}


\section{EXPERIMENTS}\label{sec5}

\subsection{Datasets and Experimental Setting}\label{sec5.1}
\emph{\textbf{Datasets.}} We evaluated our method on four long-tail benchmark datasets CIFAR-10-LT, CIFAR-100 LT \cite{paper8}, ImageNet-LT \cite{paper35}, and iNaturalist 2018 \cite{paper40}. Long-tailed CIFAR is the artificially produced imbalance dataset using its balanced version. We chose three long-tailed versions with imbalance factors (IF) of $10$, $50$, and $100$ for training. ImageNet-LT contains a total of $1000$ classes with an imbalance factor of $256$. iNaturalist 2018 is a large-scale species classification dataset with a long-tailed distribution and imbalance factor of $500$. In this work, the official training and testing splits of all datasets are used for a fair comparison.

\emph{\textbf{Experimental Setting.}} In accordance with the previous setup \cite{paper6,paper33,paper35}, we adopt ResNet-32 \cite{paper12} as the backbone network on the CIFAR-10/100-LT and adopt an SGD optimizer with momentum $0.9$ for all experiments. The batch size is set to $128$, the initial learning rate is $0.1$, and a total of $200$ epochs are trained. Linear warm-up of the learning rate is used in the first five epochs, with the learning rate decaying by $0.1$ times at $160$ and $180$ epochs respectively. We employ ResNeXt-50 \cite{paper49} on ImageNet-LT and ResNet-50 on iNaturalist 2018 as the backbone network, training $200$ epochs. In all experiments, the batch size is set to $256$ (for ImageNet-LT) / $512$ (for iNaturalist 2018), the initial learning rate is $0.1$ (linear LR decay), and the SGD optimizer with a momentum of $0.9$ is used to train all models.

\subsection{Effect of hyper-parameter $\mu$}\label{sec5.2}

\begin{figure}[h]
\vskip -0.05in
	\centering
	\begin{minipage}{0.49\linewidth}
		\centering
		\includegraphics[width=1.06\linewidth]{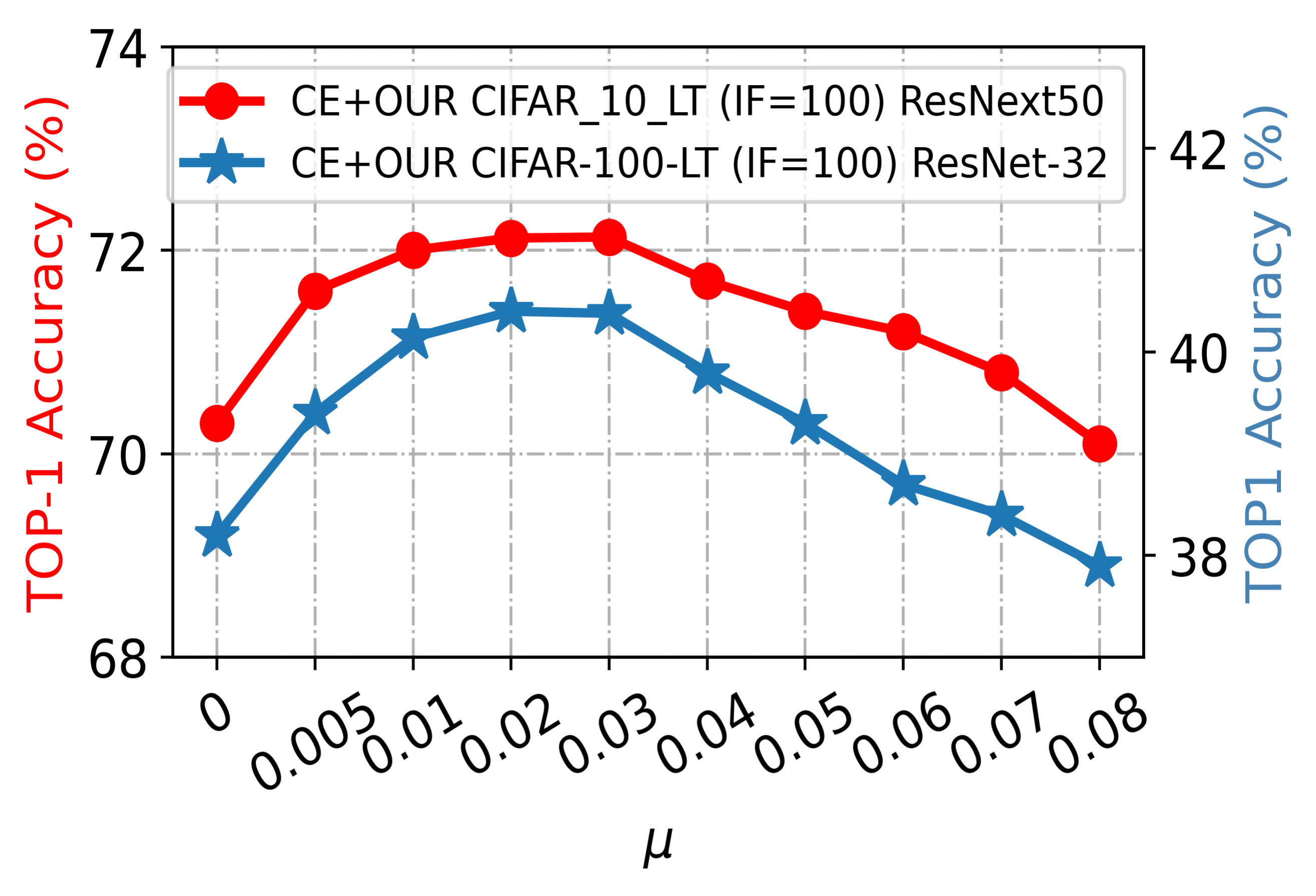}
	\end{minipage}
	\begin{minipage}{0.49\linewidth}
		\centering
		\includegraphics[width=0.96\linewidth]{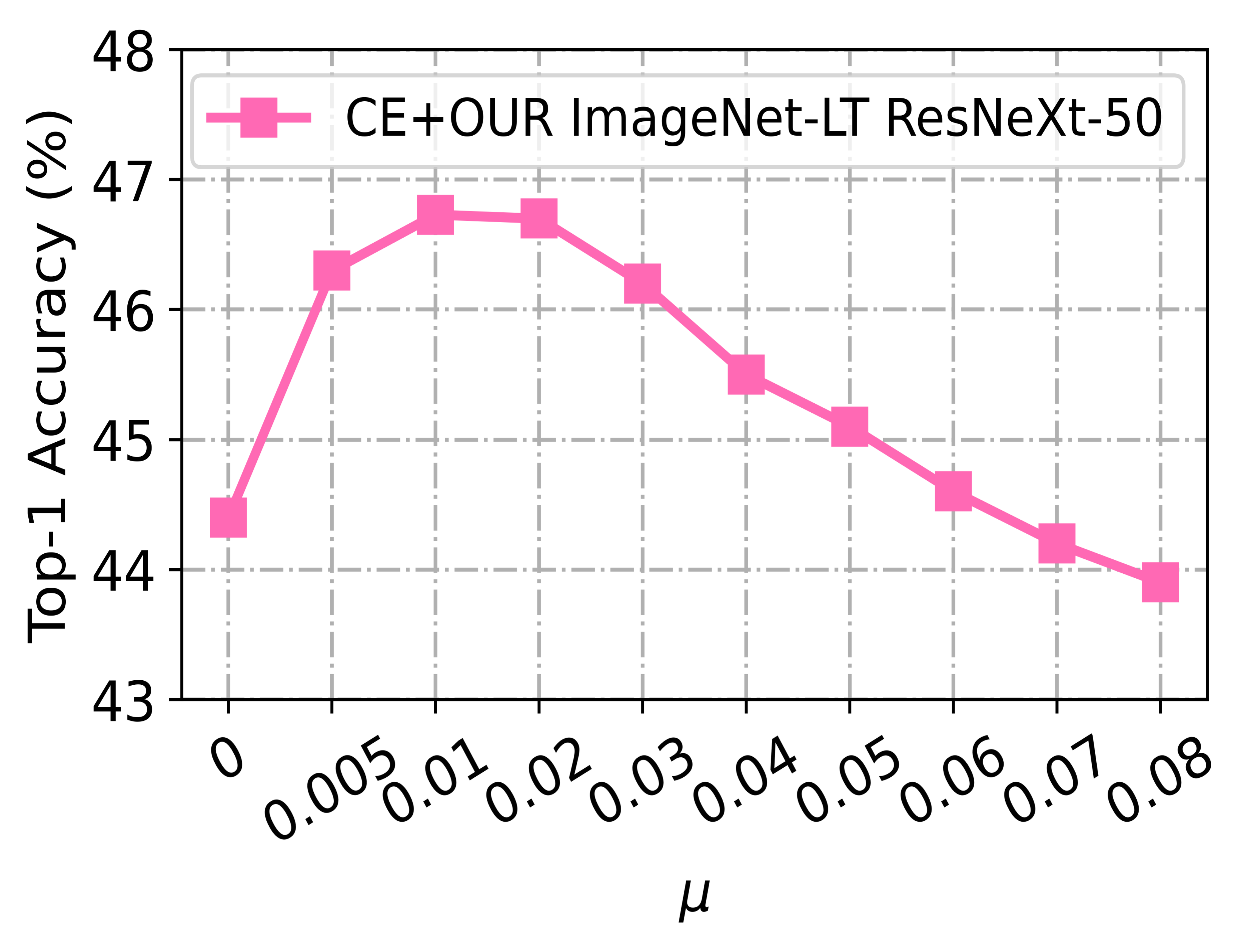}
	\end{minipage}
\vskip -0.1in
\caption{Ablation on CIFAR-10-LT (IF=$100$), CIFAR-100-LT (IF=$100$), and ImageNet-LT for select hyperparameter $\mu$.}
\label{fig5}
\vskip -0.1in
\end{figure}

$\mu$ determines the degree of uncertainty of the feature embedding, and when $\mu=0$, \emph{OUR} does not perform a transformation on the feature embedding. Since we observe that
the human eye can barely recognize samples on noisy data manifolds when $\mu=0.02$. Therefore, we explore the effect of $\mu$ on \emph{OUR} in the interval $[0,0.1]$. Experimental results on CIFAR-10-LT, CIFAR-100-LT and ImageNet-LT are shown in Fig.\ref{fig5}. 
It can be seen that the performance of the model increases and then decreases as $\mu$ increases. When $\mu$ is too small, the perturbation of the feature embedding is weak and the model does not learn sufficiently from the noise. Due to the scarcity of tail class samples, the model's learning of the original feature distribution is easily disturbed when $\mu$ is too large. Specifically, optimal performance is achieved on CIFAR-10/100-LT when $\mu$ is taken as $0.02$ and $0.03$ and on ImageNet when $\mu$ is taken as $0.01$ and $0.02$ for \emph{OUR}.

\subsection{Evaluation on CIFAR-10/100-LT}\label{sec5.3}

Table \ref{table1} shows the experimental results on CIFAR-10/100-LT. The \textcolor{pink!200}{pink rows} show improvements to the cross-entropy loss as well as cost-sensitive learning methods, and the \textcolor{blue!80}{blue rows} show improvements to the information augmentation methods. It can be observed that our proposed \emph{OUR} significantly improves the existing methods on all datasets. In particular, it also shows effectiveness on several state-of-the-art methods (RIDE+CMO, GCL, ResLT), which indicates that existing methods lack attention to the long-tailed phenomenon of robustness. It is important to note that CE is not designed for long-tailed classification, but our approach improves the performance of CE by $1.8\%$ and $1.5\%$ on CIFAR-10-LT (IF = $100$) and CIFAR-100-LT (IF = $100$) by improving the long-tailed phenomenon of model robustness alone. This further indicates that the long-tailed distribution of robustness in the long-tailed scenario limits the performance of the classification model. \emph{OUR} performs better on datasets with larger IF, which is in line with our expectation since the long tail of model robustness becomes more severe when the data are more unbalanced. For example, on the CIFAR-10-LT (IF = $100, 50, 10$) and CIFAR-100-LT (IF = $100, 50, 10$), \emph{OUR} achieved performance gains of $1.4\%$, $0.9\%$, $0.8\%$, $1.2\%$, $1.2\%$ and $0.7\%$ for CB-Focal.

\begin{table}[t]
\caption{Comparison on CIFAR-10-LT and CIFAR-100-LT. The accuracy (\%) of Top-1 is reported. The best and second-best results are shown in \underline{\textbf{underlined bold}} and \textbf{bold}, respectively.}
\label{table1}
\vskip -0.07in
\centering  
\begin{small}
\renewcommand\arraystretch{0.9}
\setlength{\tabcolsep}{6.5pt} 

\begin{tabular}{lcccccc}
\bottomrule \hline
Dataset          & \multicolumn{3}{c}{CIFAR-10-LT} & \multicolumn{3}{c}{CIFAR-100-LT} \\ \hline
Backbone Net     & \multicolumn{6}{c}{ResNet-32}                                      \\ \hline
Imbalance Factor & 100       & 50       & 10       & 100       & 50        & 10       \\ \hline
BBN \cite{paper61}              & 79.8      & 82.1     & 88.3     & 42.5      & 47.0      & 59.1     \\
De-c-TDE \cite{paper39}         & 80.6      & 83.6     & 88.5     & 44.1      & 50.3      & 59.6     \\ \hline
Cross Entropy    & 70.3      & 74.8     & 86.3     & 38.2      & 43.8      & 55.7     \\
\rowcolor{pink!25} + OUR            & 72.1      & 76.1     & 87.5     & 39.7      & 45.2      & 56.8     \\
CB-Focal \cite{paper8}         & 74.5      & 79.2     & 87.4     & 39.6      & 45.3      & 57.9     \\
\rowcolor{pink!20} + OUR            & 75.9      & 80.1     & 88.2     & 40.8      & 46.5      & 58.6     \\
LDAM-DRW \cite{paper3}         & 77.0      & 81.0     & 88.2     & 42.0      & 46.6      & 58.7     \\
\rowcolor{pink!25} + OUR            & 78.1      & 81.8     & 88.9     & 42.8      & 47.4      & 59.3     \\ \hline
MiSLAS \cite{paper60}           & 82.1      & 85.7     & \textbf{90.0}     & 47.0      & 52.3      & \textbf{63.2}     \\
\rowcolor{blue!5} + OUR            & \textbf{83.4}      & \underline{\textbf{86.7}}     & \underline{\textbf{90.8}}     & 48.1      & 53.1      & \textbf{63.9}     \\
RIDE + CMO \cite{paper33}       & -         & -        & -        & \textbf{50.0}      & 53.0      & 60.2     \\
\rowcolor{blue!5} + OUR            & -         & -        & -        & \underline{\textbf{50.8}}      & \textbf{53.9}      & 60.7     \\
GCL \cite{paper21}              & 82.7      & 85.5     & -        & 48.7      & 53.6      & -        \\
\rowcolor{blue!5} + OPeN \cite{paper53}           & 83.1      & 85.8     & -        & 49.2      & \textbf{53.9}      & -        \\
\rowcolor{blue!5} + OUR            & \underline{\textbf{83.7}}      & \textbf{86.3}     & -        & 49.8      & \underline{\textbf{54.5}}      & -        \\
ResLT \cite{paper6}            & 80.4      & 83.5     & 89.1     & 45.3      & 50.0      & 60.8     \\
\rowcolor{blue!5} + OPeN \cite{paper53}           & 80.8      & 83.8     & 89.6     & 45.8      & 50.4      & 61.2     \\
\rowcolor{blue!5} + OUR            & 81.6      & 84.3     & \textbf{90.0}     & 46.5      & 50.9      & 61.7     \\ 
\bottomrule \hline
\end{tabular}
\end{small}
\vskip -0.1in
\end{table}

We also compare \emph{OUR} with the latest noise-based augmentation method OPEN \cite{paper53}, and it can be observed that OPeN provides very weak improvements on GCL and ResLT. Random noise may cause changes in the main features of the samples, resulting in the ambiguity between samples and labels (refer to the analysis in Section \ref{sec3.1.2}). Compared with OPeN, our method not only does not add additional training samples but also has better performance.

\subsection{Evaluation on ImageNet-LT and iNat 2018}\label{sec5.4}

We report in Table \ref{table2} not only the overall performance of \emph{OUR} but also add the evaluation results of \emph{OUR} on three subsets (Head, Middle, Tail) of the two datasets. It can be observed that \emph{OUR} improves the overall performance of the other methods by at least about $1\%$ on both datasets. For the ImageNets dataset, \emph{OUR} performs superiorly, delivering performance gains of $1.6\%$, $1.4\%$, $1.2\%$ and $1.5\%$ for CE, Focal, LADE and OFA, respectively. For the iNaturalist dataset, \emph{OUR} improves the overall performance of CE, Focal and OFA by $1.5\%$, $1.4\%$ and $1.3\%$, respectively. \emph{OUR} also maintains consistent superiority in the face of the latest state-of-the-art methods. OUR improves the overall performance of RIDE+CMO by $1\%$ and $0.9\%$ on the two datasets and improves the overall performance of ResLT by $1.3\%$ and $1.1\%$ on the two datasets, respectively. It should be noted that CMO already expands the richness of tail classes by pasting the foreground of tail classes into the background of head classes, so the samples are balanced. Even in such a case, \emph{OUR} still improves the performance of RIDE+CMO, which is a solid indication that \emph{OUR} has excellent compatibility and does not conflict with the goals pursued by existing methods.

\begin{table}[t]
\caption{Comparison on ImageNet-LT and iNaturalist2018. The Top-1 Acc (\%) is reported. The best and the second-best results are shown in \underline{\textbf{underline bold}} and \textbf{bold}, respectively.}
\label{table2}
\vskip -0.08in
\centering  
\begin{small}
\renewcommand\arraystretch{0.9}
\setlength{\tabcolsep}{2.8pt} 
\begin{threeparttable}
\begin{tabular}{lcccccccc}
\hline \toprule 
\multirow{3}{*}{Methods}    & \multicolumn{4}{c}{ImageNet-LT}  & \multicolumn{4}{c}{iNaturalist 2018}  \\ \cline{2-9}
& \multicolumn{4}{c}{ResNext-50}  & \multicolumn{4}{c}{ResNet-50}  \\ \cline{2-9}
& H   &M  &T   &Overall   &H   &M   &T   &Overall \\ \hline
DisAlign \cite{paper11}  &59.9 &49.9 &31.8 &52.9 &68.0 &71.3 &69.4 &70.2  \\
MiSLAS \cite{paper1}   &65.3 &50.6 & 33.0 & 53.4 &\underline{\textbf{73.2}}  &72.4 & 70.4 & 71.6 \\ 
DiVE \cite{paper12} &64.0 & 50.4 & 31.4 & 53.1 & 70.6  & 70.0 & 67.5 & 69.1 \\  
PaCo \cite{paper13} &63.2 & 51.6 & \underline{\textbf{39.2}} & 54.4 & 69.5  & 72.3 & 73.1 & 72.3 \\ 
RIDE (3*) \cite{paper13} & 66.2 &51.7  &34.9 &54.9  & 70.2   &72.2  &72.7  &72.2 \\ 
GCL \cite{paper9} &\multicolumn{1}{c}{-}  & \multicolumn{1}{c}{-} & \multicolumn{1}{c}{-} & 54.9 &\multicolumn{1}{c}{-}  & \multicolumn{1}{c}{-} & \multicolumn{1}{c}{-} &72.0  \\ \hline

CE &65.9 & 37.5 & 7.70 & 44.4 & 67.2  & 63.0 & 56.2 & 61.7 \\
\rowcolor{pink!20} + OUR &65.0 &38.4 & 14.5 & 46.0 & 67.3  & 63.9 & 60.5 & 63.2 \\ 

Focal Loss \cite{paper3} &\textbf{67.0} & 41.0 & 13.1 & 47.2 & \multicolumn{1}{c}{-}  & \multicolumn{1}{c}{-} & \multicolumn{1}{c}{-} & 61.1 \\
\rowcolor{pink!20} + OUR  &\underline{\textbf{67.2}} & 42.5 &19.7 &48.6 &68.6  &63.4 &57.8 &62.5 \\ 

LDAM \cite{paper2} &60.0 & 49.2 & 31.9 & 51.1 & \multicolumn{1}{c}{-}  & \multicolumn{1}{c}{-} & \multicolumn{1}{c}{-} & 64.6\\
\rowcolor{pink!20} + OUR &60.6 & 50.0 &33.5 &52.2 & 69.0  &66.9  &62.1 &65.5 \\ 

LADE \cite{paper14} &62.3 & 49.3 & 31.2 & 51.9 & \multicolumn{1}{c}{-}  & \multicolumn{1}{c}{-} & \multicolumn{1}{c}{-} & 69.7\\
\rowcolor{pink!20} + OUR &62.4 & 50.5 &34.4 &53.1 &72.2  &70.6  &65.9  &70.7 \\ \hline

OFA \cite{paper10}    &47.3 & 31.6 & 14.7 & 35.2 & \multicolumn{1}{c}{-}  & \multicolumn{1}{c}{-} & \multicolumn{1}{c}{-} & 65.9 \\
\rowcolor{blue!5} + OUR &47.2 &32.8 &18.6 &36.7 &69.7  &68.2  &64.8  &67.2 \\ 

RIDE + CMO \cite{paper8}  &66.4 & 54.9 & 35.8 & 56.2 &70.7  & 72.6 &73.4 & 72.8 \\ 
\rowcolor{blue!5} + OPeN      &66.7 & \textbf{55.1} &37.0 & \textbf{56.8} &70.4  &\textbf{73.4} &\underline{\textbf{74.1}} &\underline{\textbf{73.2}} \\
\rowcolor{blue!5} + CR      &66.5 & \underline{\textbf{55.7}} &\textbf{37.9} &\underline{\textbf{57.2}} &70.5  &\textbf{73.9} &\underline{\textbf{74.8}} &\underline{\textbf{73.7}} \\

ResLT \cite{paper15} &63.0 & 50.5 &  35.5 & 53.0 & 68.5  &  69.9 &70.4 & 70.2  \\
\rowcolor{blue!5} + OPeN      &63.3 &51.3 &36.2 & 53.6 &68.6  &70.5 &71.2 &70.7 \\
\rowcolor{blue!5} + OUR &63.5 &51.7 &37.3 &54.3 &68.8 &71.1 &72.0 & 71.3 \\ 

\bottomrule \hline
 \end{tabular}
\begin{tablenotes}
\item \footnotesize \emph{RIDE ($3$*) denotes the RIDE model with $3$ experts. RIDE in RIDE+CMO comes with $3$ experts. H, M, and T denote the Head (more than $100$ images), Middle ($20$-$100$ images), and Tail (less than $20$ images) subsets of the dataset, respectively.}
\end{tablenotes}
\end{threeparttable}
 \end{small}
 \vskip -0.15in
 \end{table}

\emph{OUR} delivers the most significant improvement for the tail subset. On ImageNet-LT, \emph{OUR} improves the performance of CE, Focal, and OFA on the tail subset by 7.5\%, 6.6\%, and 4.6\%, respectively. Compared to OPeN, our method results in better performance of RIDE+CMO and ResLT on both datasets. Specifically, RIDE+CMO+\emph{OUR} outperforms RIDE+CMO with OPeN by 0.9\% and 0.7\%, respectively, on the tail subset of both datasets, and ResLT+\emph{OUR} outperforms ResLT+OPeN by 1.1\% and 0.8\%, respectively. Comprehensive experiments on CIFAR-10/100-LT, ImageNet-LT, and iNaturalist 2018 show that our method is stable and excellent, outperforming the recently advanced noise-based augmentation method OPEN. The experiments and analysis in Section \ref{sec5.6} further demonstrate the performance gap between OPEN and \emph{OUR}.

\subsection{Evaluation with multiple backbones}\label{sec5.5}

To fully evaluate the generality of \emph{OUR}, we adopt different backbone networks on ImageNet-LT to demonstrate the effective improvement of \emph{OUR} for other methods. The experimental results are shown in Table \ref{table3}, with the \textcolor{cyan}{cyan rows} illustrating the method with ResNet-18 as the backbone network and the \textcolor{violet}{violet rows} illustrating the method with ResNeXt-101-32$\times$4d as the backbone network. \emph{OUR} brings about a $1\%$ performance gain in overall accuracy for all methods, with CE+\emph{OUR} outperforming CE by $1.2\%$ overall when ResNet-18 is used as the backbone network. Consistent with Table \ref{table2}, \emph{OUR} has the most significant performance in tail classes. For example, when ResNet-10 is used as the backbone, CE+\emph{OUR} outperforms CE by $2.4\%$ on the tail subset. When ResNeXt-101-32$\times$4d is adopted as the backbone, cRT+\emph{OUR} outperforms cRT by $1.8\%$ on the tail subset. The evaluation of various backbones solidly demonstrates that our method can work in a wide range of scenarios.

\begin{table}[t]
\caption{Top-1 Accuracy (\%) with Various ResNet Backbones.}
\label{table3}
\vskip -0.07in
\centering  
\begin{small}
\renewcommand\arraystretch{0.93}
\setlength{\tabcolsep}{5.4pt} 
\begin{tabular}{llcccc}
\hline \toprule 
Methods & Backbone Net & Head &Middle & Tail & Overall \\ \midrule
CE      & ResNet-10    & 59.7 &29.4   & 5.7  & 37.3    \\ 
\rowcolor{cyan!10} +OUR    & ResNet-10    & 60.0 & 30.3                        & 8.1(+2.4)  & 38.5(+1.2)    \\
cRT     & ResNet-10    & 53.8 & 41.3                        & 25.4 & 43.2    \\
\rowcolor{cyan!10} +OUR    & ResNet-10    & 54.0 & 42.2                        & 26.7(+1.3) & 44.0(+0.8)    \\
LWS     & ResNet-10    & 51.8 & 42.2                        & 28.1 & 43.4    \\
\rowcolor{cyan!10} +OUR    & ResNet-10    & 52.4 & 42.5                        & 29.2(+1.1) & 44.1(+0.7)    \\
ResLT   & ResNet-10    & 52.3 & 41.6                        & 27.6 & 43.0    \\
\rowcolor{cyan!10} +OUR    & ResNet-10    & 52.7 & 42.5                        & 29.0(+1.4) & 43.9(+0.9)    \\ \midrule
CE      & ResNeXt-101  & 69.6 & 44.6                        & 15.6 & 49.6    \\
\rowcolor{violet!10} +OUR    & ResNeXt-101  & 69.9 & 45.7                        & 17.1(+1.5) & 50.6(+1.0)    \\
cRT     & ResNeXt-101  & 66.2 & 50.4                        & 30.8 & 53.3    \\
\rowcolor{violet!10} +OUR    & ResNeXt-101  & 66.4 & 51.7                        & 32.6(+1.8) & 54.4(+1.1)    \\
LWS     & ResNeXt-101  & 65.7 & 51.4                        & 34.7 & 54.0    \\
\rowcolor{violet!10} +OUR    & ResNeXt-101  & 66.0 & 52.0                        & 36.3(+1.6) & 54.8(+0.8)    \\
ResLT   & ResNeXt-101  & 63.3 & 53.3                        & 40.3 & 55.1    \\
\rowcolor{violet!10} +OUR    & ResNeXt-101  & 63.8 & 54.1                        & 41.7(+1.4) & 56.0(+0.9)   \\
\bottomrule \hline
\end{tabular}
\end{small}
\vskip -0.1in
\end{table}

\subsection{CONCLUSION}\label{sec6}

This work finds and defines the long-tail phenomenon of model robustness in data imbalance scenarios and proposes a corresponding calculation of the imbalance factor (i.e., \emph{RIF}). Then, we propose the orthogonal uncertainty representation (\emph{OUR}) of features to mitigate the model bias on data manifolds and noisy data manifolds. Also, an end-to-end training scheme is proposed for efficient and fast application of \emph{OUR}. Although our approach strongly mitigates the degree of imbalance in model robustness, it still needs to be improved. In the future, we hope more work will focus on the long-tail phenomenon of model robustness to make the model more robust in data imbalance scenarios.

\begin{acks}
\small This work was supported in part by
the Key Scientific Technological Innovation Research Project by Ministry of Education,
the State Key Program and the Foundation for Innovative Research Groups of the National Natural Science Foundation of China (61836009),
the Major Research Plan of the National Natural Science Foundation of China (91438201, 91438103, and 91838303),
the National Natural Science Foundation of China (U22B2054, U1701267, 62076192, 62006177, 61902298, 61573267, 61906150, and 62276199),
the 111 Project,
the Program for Cheung Kong Scholars and Innovative Research Team in University (IRT 15R53),
the ST Innovation Project from the Chinese Ministry of Education,
the Key Research and Development Program in Shaanxi Province of China(2019ZDLGY03-06),
the National Science Basic Research Plan in Shaanxi Province of China(2022JQ-607),
the China Postdoctoral fund(2022T150506),
the Scientific Research Project of Education Department In Shaanxi Province of China (No.20JY023),
the National Natural Science Foundation of China (No. 61977052).
\end{acks}

\bibliographystyle{ACM-Reference-Format}
\balance
\bibliography{sample-base}



\newpage
\appendix

\section{OUR mitigates the long tail of model robustness}\label{sec5.6}

\begin{figure}[t]
	\centering
	\begin{minipage}{0.495\linewidth}
		\centering
		\includegraphics[width=1\linewidth]{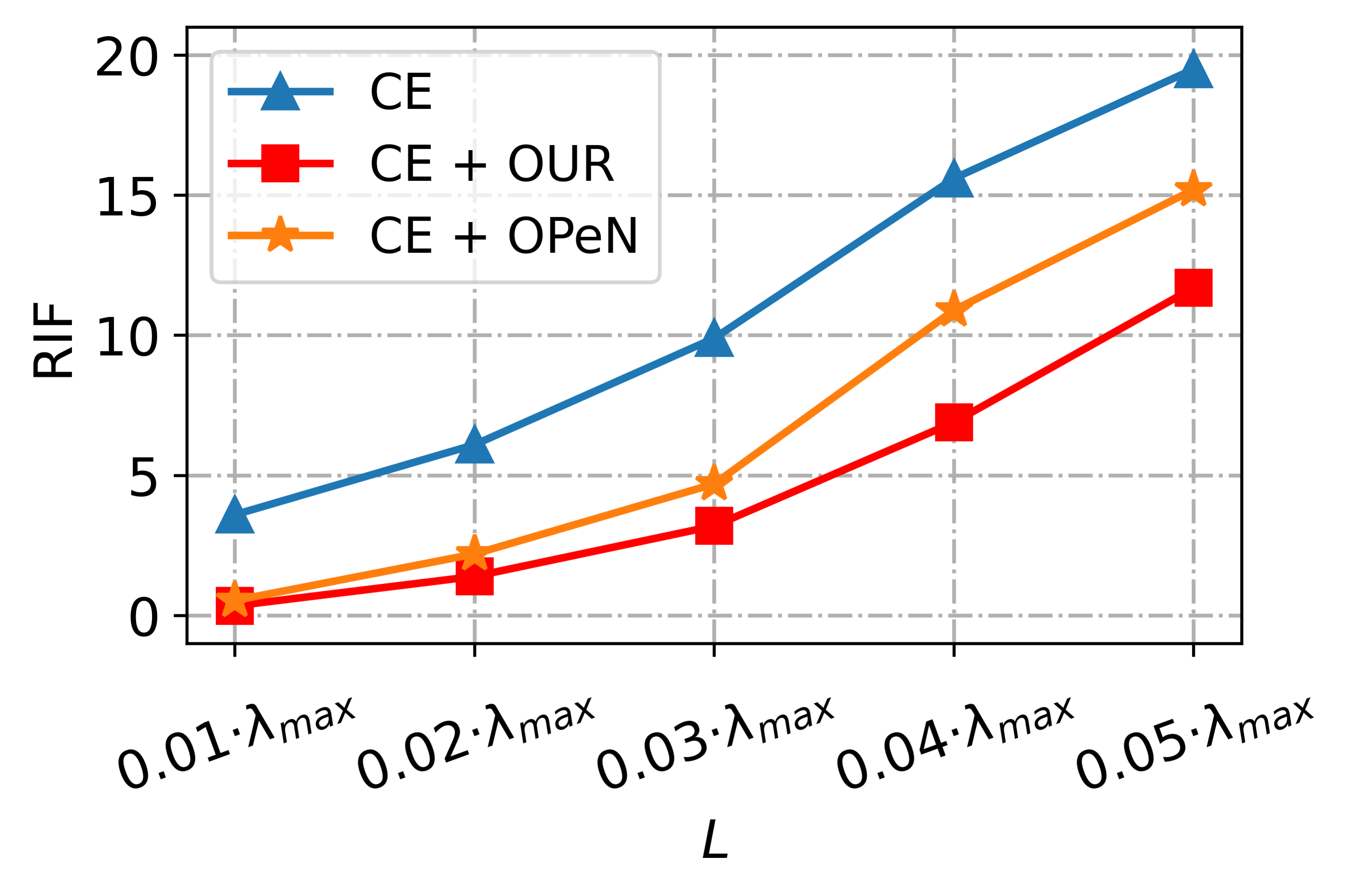}
	\end{minipage}
	\begin{minipage}{0.495\linewidth}
		\centering
		\includegraphics[width=1\linewidth]{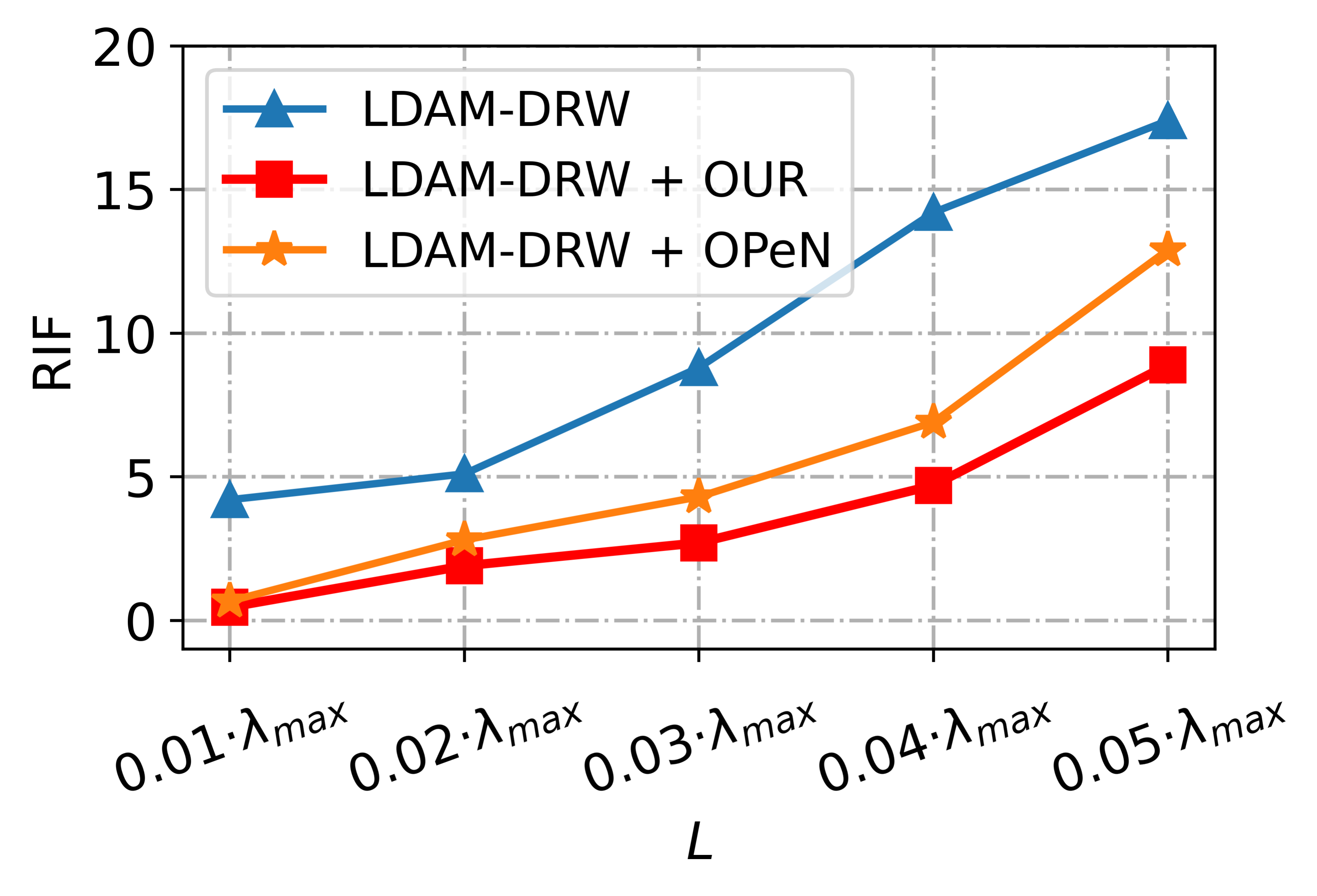}
	\end{minipage}

	\begin{minipage}{0.495\linewidth}
		\centering
		\includegraphics[width=1\linewidth]{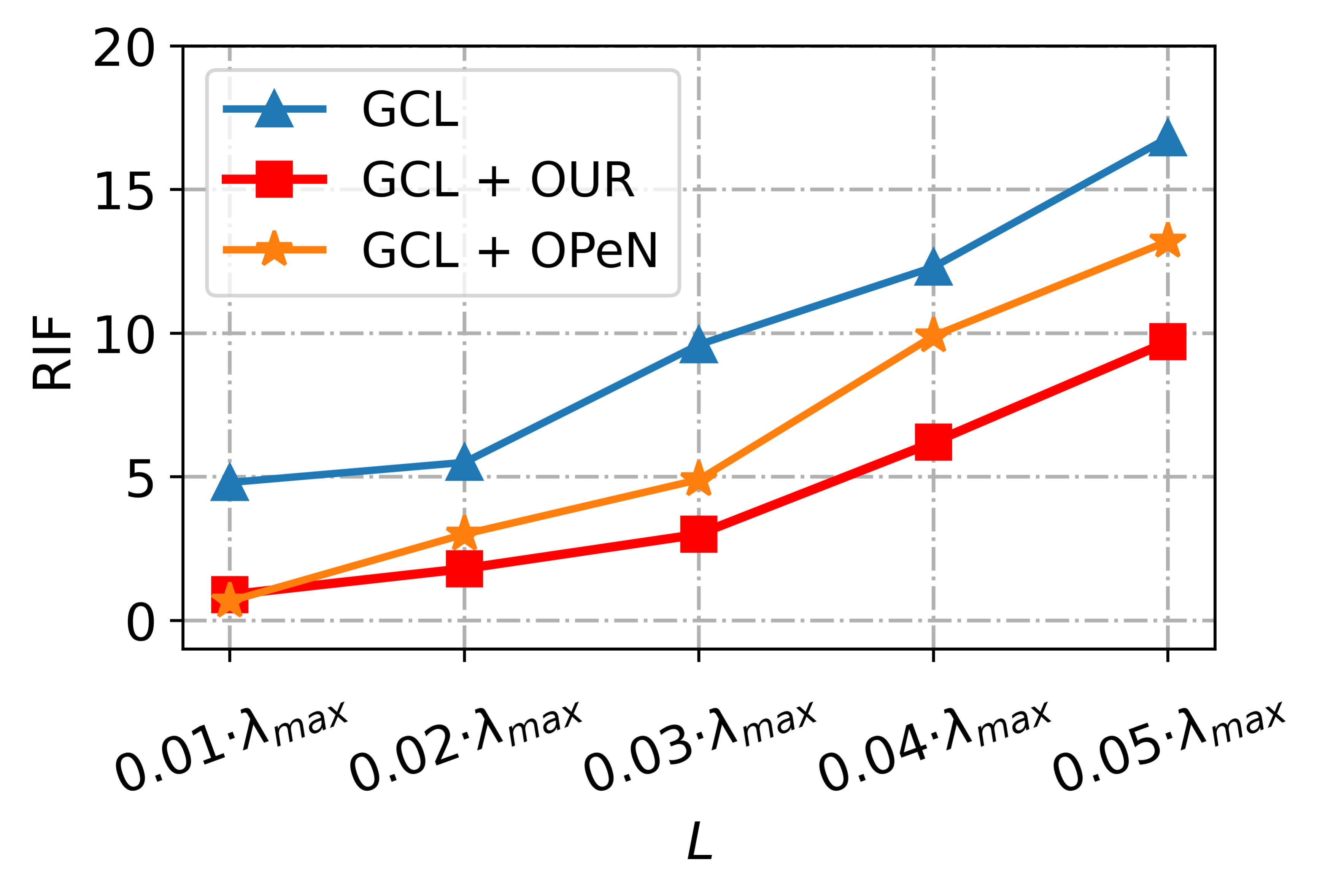}
	\end{minipage}
	\begin{minipage}{0.495\linewidth}
		\centering
		\includegraphics[width=1\linewidth]{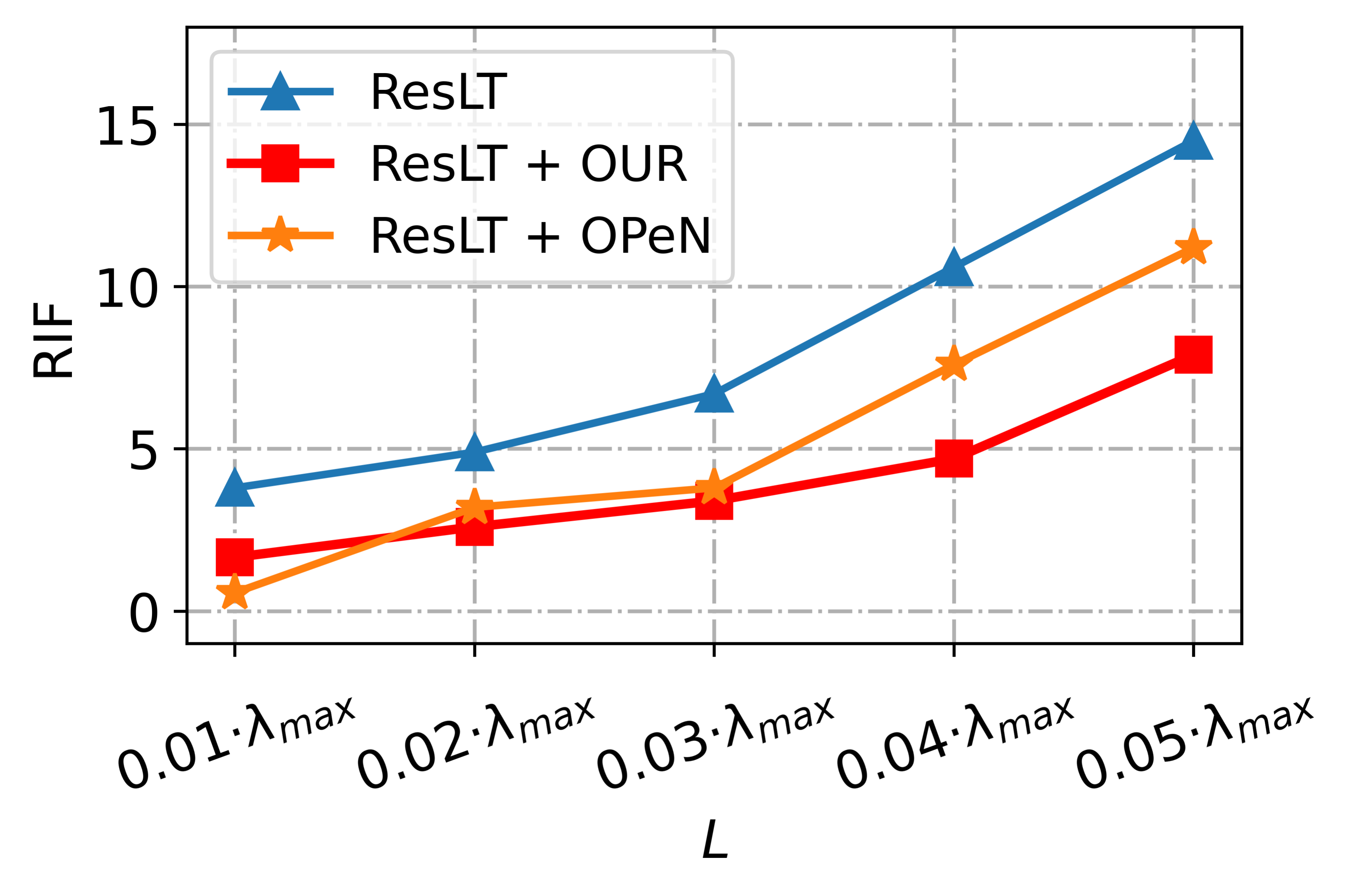}
	\end{minipage}
\vskip -0.05in
\caption{OUR alleviates the imbalance of model robustness. $L$ denotes the distance between the noisy data manifold and the data manifold.}
\label{fig6}
\vskip -0.05in
\end{figure}

To visualize the improvement effect of \emph{OUR} on the long-tailed phenomenon of robustness, we constructed multiple noisy data manifolds corresponding to the training data of CIFAR-10-LT (IF = 100), and then calculate the values of \emph{RIF} (Definition 2) before and after using \emph{OUR} for CE, LDAM-DRW, GCL, and ResLT. Experimental results are illustrated in Fig.\ref{fig6}. The imbalance degree \emph{RIF} of model robustness is significantly reduced by adopting \emph{OUR} to improve multiple methods, and the ability of OPeN to mitigate the long tailed phenomenon of robustness is between the original method and \emph{OUR}. When $L$ is small, the improvement effect of OPeN on \emph{RIF} is close to \emph{OUR}, but as $L$ increases, the effect of OPeN becomes weak. This may be due to the fact that OPEN is based on pure noise and its randomness of direction leads to the inability to generate noisy data manifolds stably at long distances, which limits the performance.

\end{document}